\documentclass[12pt]{article}
\usepackage[margin=1in]{geometry}
\usepackage{graphicx}
\usepackage{float}

\usepackage[authordate,bibencoding=auto,strict,backend=biber,natbib]{biblatex-chicago}
\begin{document}

\title{AIGC In China: Current Developments And Future Outlook}
\date{}
\author{Xiangyu Li$^{1}$, Yuqing Fan$^{2,3}$, Shenghui Cheng$^{2,3}$\\
$^1$ School of Shandong Foreign Languages Vocational College, Rizhao, China\\
$^2$ Research Center for the Industries of the Future, \\ Westlake University, Hangzhou, China \\
$^3$ School of Engineering, Westlake University, Hangzhou, China,\\
}

\maketitle 

\begin{abstract}

The increasing attention given to AI Generated Content (AIGC) has brought a profound impact on various aspects of daily life, industrial manufacturing, and the academic sector. Recognizing the global trends and competitiveness in AIGC development, this study aims to analyze China's current status in the field. The investigation begins with an overview of the foundational technologies and current applications of AIGC. Subsequently, the study delves into the market status, policy landscape, and development trajectory of AIGC in China, utilizing keyword searches to identify relevant scholarly papers. Furthermore,  the paper provides a comprehensive examination of AIGC products and their corresponding ecosystem, emphasizing the ecological construction of AIGC. Finally, this paper discusses the challenges and risks faced by the AIGC industry while presenting a forward-looking perspective on the industry's future based on competitive insights in AIGC.

\end{abstract}

\section{Introduction}

AIGC, or Artificial Intelligence Generated Content, refers to the production of content facilitated by artificial intelligence technologies. The introduction of ChatGPT has significantly impacted various aspects of our lives and professional endeavors. It utilizes input data and existing information to conduct insightful analyses and generate relevant content based on user preferences. For example, AIGC can offer article suggestions and assistance in writing. Similarly, it can aid visual artists by providing relevant images, while video and music creators can benefit from its ability to generate video and audio content. In terms of everyday life, ChatGPT can serve as a valuable resource for finding solutions, such as recipes for cooking, ideas for children's crafts, and guidance on setting up smart homes. In professional settings, ChatGPT can streamline tasks such as data organization, report generation, and even assist in creating presentations, programming, and website development. By reducing time spent on routine tasks and enhancing production efficiency, AIGC enables individuals to allocate more time to personal commitments, family engagements, and fostering stronger relationships within their professional teams. 

AIGC has emerged as a disruptive and transformative technology, offering significant advantages in promoting economic growth and enhancing productivity. Its potential to reshape the global industry is widely recognized, leading countries like China to prioritize the development of artificial intelligence within their national agendas. This paper aims to explore the current state of AIGC development in China and assess its impact on the country's global competitiveness in this field.

Currently, AIGC has garnered significant societal attention in China, primarily due to the proliferation of large-scale models and open-source frameworks. This development has positioned AIGC as a promising field for the practical application of artificial intelligence. On one hand, the accessibility and availability of large models and open-source frameworks have lowered the barriers to entry for AIGC and expanded its application potential. On the other hand, the boundaries between AI and innovation have become increasingly blurred, highlighting the growing trend of their integration (\cite{ref101}).

To draw a conclusion, AIGC constitutes a profound transformation that will significantly impact China's social, technological and economic lives. This comprehensive article aims to provide a detailed introduction to the concept of current development in China. The contributions of this paper including: 1) a thorough exposition of the compositions and applications of AIGC, and a comprehensive analysis of its current development status from the perspectives of policy, market, and scientific research in China; 2) an exploration of the changes and opportunities of AIGC, identifying its developmental requirements based on these changes and opportunities; and 3) an overview of the future prospects for the evolution of AIGC in China.
 
\section{Compositions and applications}

This section first introduces the foundation technology of AIGC, then provides some common applications that has been used widely.

\subsection{Foundations of AIGC}

The development of AIGC has been ongoing for a considerable period of time. Its success today can be attributed to the exploration of several other computer technologies. Many scholars have reviewed the foundational technologies of AIGC. For example, \cite{cao2023comprehensive} identified Foundation models, Reinforcement learning with human feedback (RLHF), and advancements in computing technology (such as hardware, distributed computing, and cloud computing) as AIGC key foundational technologies. \cite{zhang2023complete} categorized the founding technology of AIGC into two groups: general techniques and creation techniques. General techniques include Backbone architecture and Self-supervised pretraining, while creation techniques encompass likelihood-based models, energy-based models, GAN, and the Transformation model. \cite{wu2023aigenerated} argued that the development of AIGC technology requires three essential elements: data for training models, hardware for computing power, and algorithms for model establishment. They also proposed a five-stage evolution of AI models, progressing from Machine Learning, Neural Network Model (NNM), Generative algorithm, Language Model, to Large-scale Pre-trained Model. 

We suggest that the foundations of AIGC technology lie in Large Language Models (LLM), pre-trained models, Multimodal Models, and Reinforcement learning with human feedback.

\subsubsection{Large Language Models}

A large language model (LLM) is a neural network-based system employed in natural language processing to comprehend the underlying patterns and structures of human language. These models are trained through unsupervised learning techniques and have the ability to generate coherent and grammatically accurate sentences without explicit programming or specific rule sets.

The Transformer model is widely used in contemporary deep learning models, it represents a significant advancement in large language models. This methodology was developed to overcome the limitations of traditional Recurrent Neural Networks in effectively processing variable-length input sequences while maintaining contextual awareness. Notably, the Transformer model incorporates a self-attention mechanism, which allows the model to selectively focus on and weigh different segments of the input sequence (\cite{ref2201}).

\begin{figure}[H]
    \centering
    \includegraphics[scale=0.7]{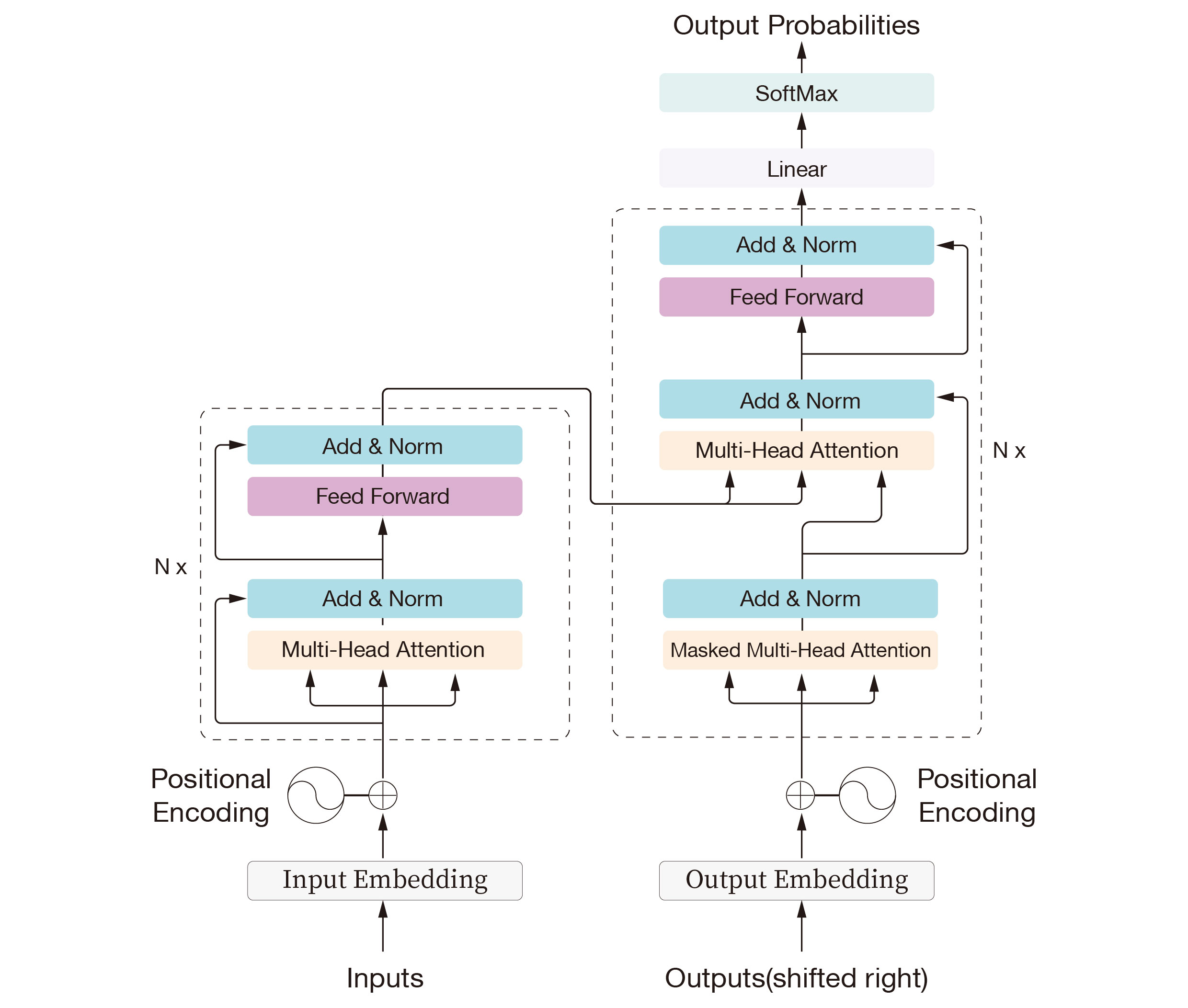} 
    \caption{Structure of Transformer}
    \label{transformer}
\end{figure}

The Transformer model consists of two primary components: an encoder and a decoder. The encoder operates on the input sequence, generating hidden representations, while the decoder utilizes these representations to produce an output sequence. Each layer of the encoder and decoder comprises a multi-head attention sub-component combined with a feed-forward neural network. The core functionality of the multi-head attention module allows for assigning varying weights to individual tokens based on their relevance. This capability enhances the model's performance across a wide range of Natural Language Processing (NLP) tasks and improves its ability to handle long-term dependencies, particularly those with intricate syntactic structures (\cite{vaswani17}). Figure \ref{transformer} illustrates the structure of the transformer model.

In addition to these advantages, a notable characteristic of the Transformer model is its heightened parallelizability and capacity to overcome inherent biases, thus enhancing its viability for extensive pre-training tasks. These pre-training tasks facilitate the creation of versatile transformer-based models that can be applied effectively to a diverse range of downstream problems. As a result, the Transformer has emerged as a critical component in contemporary natural language processing applications, representing the state-of-the-art in the field (\cite{cao2023comprehensive}).

\subsubsection{Pre-trained models}

To address the challenge of word ambiguity, scholars proposed integrating word vector models into subsequent task models during the training process. This concept laid the foundation for pre-trained models, with the ElMo model utilizing Bi-LSTM serving as the first successful case (\cite{wawer21}). Subsequently, the renowned GPT model emerged, representing a typical pre-trained language model designed to comprehend text effectively and subsequently fine-tuned for various downstream tasks, embodying the concept of a single model solving multiple problems.

GPT-1, the initial model in the GPT series, is a pre-trained language model constructed by stacking 12 Transformer decoder models. Like other pre-trained language models, GPT-1 undergoes two phases to carry out specific tasks: pre-training and fine-tuning. The objective of the pre-training phase is to equip the GPT-1 model with an understanding of the text's content by implementing appropriate pre-training tasks. Specifically, the pre-training task can be perceived as a text completion task. Initially, a vast corpus comprising high-quality natural language texts is collected, and each text is partially masked. The model reads the text before masking and predicts the masked content one token at a time. Following pre-training, the GPT-1 model can be fine-tuned for specific downstream tasks. For instance, in text classification, the training texts are annotated accordingly, such as labeling positive as 0 and negative as 1. Each training text is then manually annotated, marking this form of training as supervised training (\cite{alec18}).

Subsequently, researchers further explored the potential of GPT-1 and put forth a daring hypothesis: if the GPT model's understanding of a text is further enhanced, can it generate reasonable answers solely based on textual input? Based on this hypothesis, scholars advanced the GPT-2 model (\cite{wei21}). The core idea behind GPT-2 is to eliminate the fine-tuning process utilized in GPT-1 and employ the question itself as the input after pre-training. This approach enables the model to directly generate answers through text generation. This input type is often referred to as a "prompt." For tasks like text sentiment classification, determining the sentiment of a statement can be accomplished by designing a suitable prompt that frames the specific question. The text generation model can generate positive or negative directly, thereby completing the text classification. In such cases, the prompt does not provide any explicit indications of how the question should be answered, a phenomenon known as zero-shot learning.

GPT-3 can be viewed as an enhanced version of GPT-2 (\cite{brown20}). The structural difference between the two models is insignificant. However, the parameter size of GPT-3 has further increased from 1.5 billion in GPT-2 to 175 billion. Another noteworthy innovation in GPT-3 lies in the design of prompts. Researchers, inspired by GPT-2's ability to accomplish downstream tasks without fine-tuning, propose incorporating specific hints in prompt design to enhance the model's performance for particular tasks. More specifically, GPT-3's training data no longer comprises solely natural language text but includes high-quality prompts tailored for specific tasks. Each prompt encompasses several small example prompts. For instance, in text sentiment translation, the corresponding prompt can be devised with numerous small samples. With the assistance of these slight prompts, the model can obtain the desired answer through forward computation alone, a phenomenon known as few-shot learning. Through the implementation of few-shot learning, GPT-3 has showcased the potential of this learning approach to the wider scientific community.

\subsubsection{Multimodal Models}

Multimodal models encompass models that integrate heterogeneous data from different modalities to perform a variety of tasks. Each modality presents distinct data types that can be utilized collectively in a task. By incorporating diversified data sources, multimodal models acquire richer information, thereby improving the accuracy and efficiency of tasks (\cite{baltru18}). In this section, we introduce two prominent multimodal models prevalent in contemporary research: CLIP and DALL-E.

CLIP (Contrastive Language-Image Pre-Training) is a neural network-based model devised and developed by OpenAI, aimed at training representations for both textual and visual data. This model employs a standard training approach to simultaneously train representations for text and images, leading to improved performance across these two domains. The core concept of the CLIP model involves mapping images and text into a standardized vector space, wherein similar images and text are positioned closer to each other within this space. This vector space construction method, known as contrastive learning, harnesses the concepts of similarity and dissimilarity to train the model by minimizing the distance between data belonging to the same class and maximizing the distance between data from different classes (\cite{alec21}).

DALL-E is a neural network-powered image generation model developed by OpenAI, designed to generate images aligned with a given textual description. The fundamental principle behind DALL-E is grounded in technologies such as Generative Adversarial Networks (GANs) and Variational Autoencoders (VAEs), which enable the conversion of text descriptions into corresponding images and the generation of images consistent with the input text. Specifically, DALL-E comprises two primary components: an image generation network and a text encoder. The text encoder transforms the provided textual description into a vector, which is subsequently fed into the image generation network. This network leverages the vector representation to generate an image that accurately corresponds to the given description. DALL-E employs large-scale self-supervised learning to autonomously acquire knowledge from internet-sourced images, empowering it to generate novel objects that have not been previously encountered. Furthermore, DALL-E incorporates specialized techniques such as multi-layered skinning, inlaying, and continuous geometry to enhance the realism and diversity exhibited by the generated images (\cite{ref307}).

\subsubsection{RLHF}

The success of ChatGPT can be attributed to its utilization of reinforcement learning from human feedback (RLHF) in combination with the GPT-3 model, as well as the implementation of a harm reduction approach to generating answers (\cite{lee23}). To elaborate, RLHF involves training a reward model (RM) to simulate human language preferences and employing reinforcement learning to train the language model accordingly, aiming to generate responses that increasingly align with human preferences (\cite{ouyang22}). 

The RLHF process consists of three steps. Initially, a set of approximately 10,000 prompts from past user data is selected, covering a diverse range of task types and topics, while high-quality responses are annotated as labels. Next, a feedback model is trained to evaluate language model-generated responses using scores indicative of human preferences. Different answers for each prompt are gathered and annotated personnel rank them based on quality, forming the learning objective for training the feedback model. Lastly, Proximal Policy Optimization (PPO), an optimization reinforcement learning algorithm, is employed by developers to further enhance the performance of the model. This algorithm enables the dynamic adjustment of the language model's strategy based on the feedback model's scores, with the aim of achieving higher scores. As the feedback model assigns higher scores to responses more favorably received, the training process ensures that the model generates responses that increasingly align with human preferences (\cite{john17}).

\subsection{Applications of AIGC}

\begin{figure}[H]
    \centering
    \includegraphics[scale=0.7]{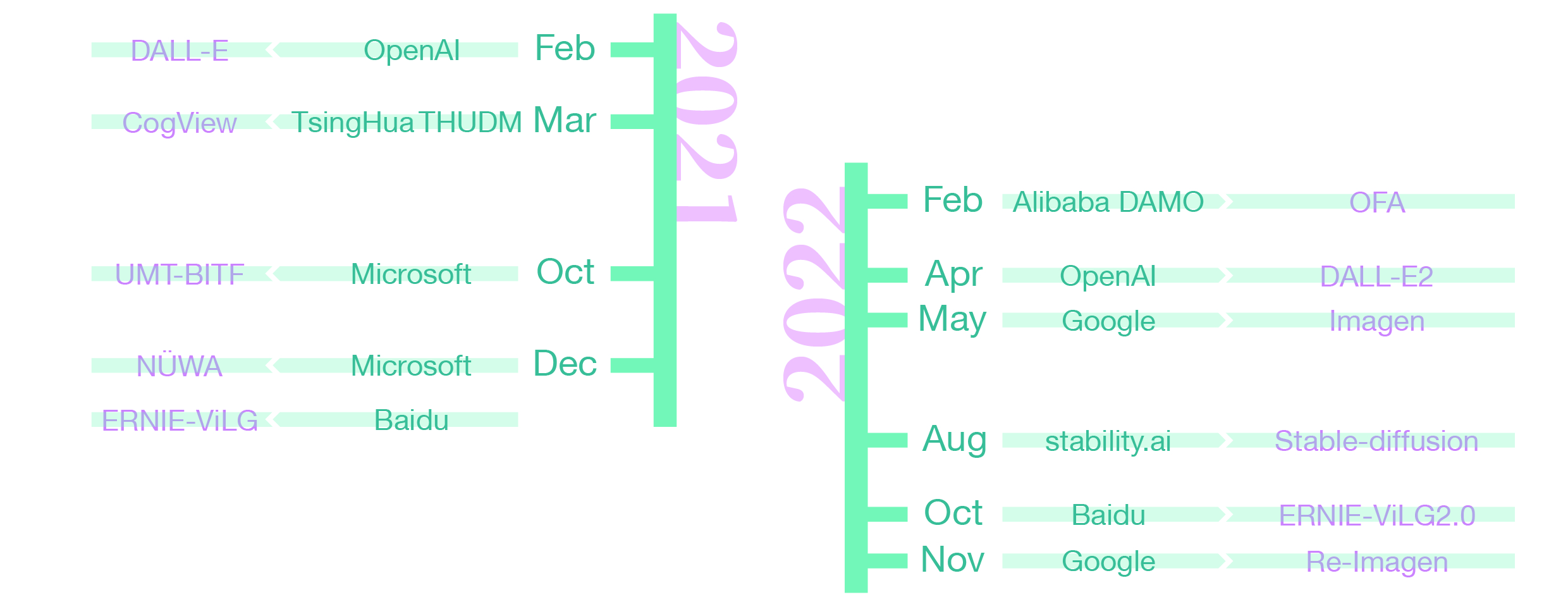} 
    \caption{Several famous generative models after 2021}
    \label{models}
\end{figure}

The usage of AIGC in text and image generation has become increasingly mature. Since 2021, many leading internet companies at home and abroad have released various models for generating text and images. These text and image models have achieved significant results in various application areas and have released many software and hardware products for users. Figure \ref{models} presents several generative models after 2021.

\subsubsection{Text generating}

AIGC plays a crucial role in facilitating digital transformation as it offers the ability to automate and expedite content creation processes. Through the utilization of AI, high-quality text, images, audio, or video content can be generated based on specified inputs and requirements. This feature significantly assists individuals and businesses in swiftly producing diverse content, including product descriptions, news releases, and social media posts. The automation of content creation not only leads to substantial time and labor cost savings but also enhances the efficiency and scalability of content production.

The prevailing learning approach begins by establishing the learning style using algorithms. Subsequently, the AI learns from vast amounts of textual data and utilizes deep learning models and existing learning outcomes to generate novel text and content. The following outlines the six training steps employed in AI content generation: training, encoding, learning, generation, optimization, and iteration.

In addition to its usage in generating articles, AIGC is also highly applicable in code generation. Users can input abstract descriptions or objectives, and AI can automatically generate corresponding code. Compared to conventional coding methods, AIGC alleviates the workload of developers and enhances development efficiency, which allows developers to spend time and effort on more important issues.

\subsubsection{Picture generating}

The utilization of AI for the creation of artistic content has witnessed significant growth. These techniques involve training algorithms on extensive datasets of pre-existing artwork. Through machine learning, these methods are capable of producing new artwork that either simulates the styles and techniques of renowned artists or develops novel artistic styles altogether. The remarkable advancements in diffusion-based models have prompted numerous companies to introduce a multitude of AI-generated art products.

Among these models, DreamStudio, engineered by Stability.ai, stands out. This platform utilizes stable diffusion techniques to generate images based on provided phrases or sentences. DreamStudio exhibits comparable performance to DALL-E-2, while boasting faster processing speeds (\cite{Rombach_2022_CVPR}).

State-of-the-art AI software integrated with text-to-image algorithms empowers the generation of portraits, landscapes, abstract art, and even the replication of famous artists' styles within minutes. For instance, Anges Russel from Sydney established a web-based platform called Night Cafe Creator in 2019. Anges employed a novel model named VQGAN+CLIP which demonstrated impressive results in creating oil painting-style images. Due to its success, the Night Cafe application rapidly gained popularity, solidifying its position as one of the most sought-after image generation software applications (\cite{nightcafe}).

\subsubsection{Other applications}

In addition to previous applications, AIGC has made significant advancements in various other domains. AI Text-to-Speech (TTS) converts written text into natural and fluent speech. It has the ability to transform input text into audible speech, replicating human speech features and expressions. Applications of AI-TTS such as Deep Voice 3 had experienced significant growth and is widely utilized in fields like voice assistants, voice navigation, and voice broadcasting, enabling the production of high-quality and lifelike speech output (\cite{ping2018deep}).

AI has also been applied to music editing and composition, serving as a tool to aid or automate the process. Music producers, composers, and enthusiasts can benefit from various AI-powered tools and systems that provide creative inspiration and assistance during music creation and editing.

Furthermore, AIGC has the potential to revolutionize the gaming industry by introducing innovative gameplay. With AI strategy generation, virtual characters, known as Non-Player Characters (NPCs), can possess intelligent decision-making abilities and behavior within games. Unlike conventional NPCs that rely on predefined rules and scripts, AI-based NPCs can autonomously learn from analyzing the game environment and player actions, allowing them to dynamically adjust their strategies and provide more challenging and realistic gaming experiences (\cite{ARZATECRUZ201711}).

\section{Development status}

The inception of AIGC can be traced back to the 1980s, when researchers initiated investigations into machine learning, natural language processing, and related technologies for text content generation. In the 1990s, the emergence of deep learning algorithms prompted researchers to explore the utilization of neural networks for text generation. However, limitations in hardware capabilities and data availability hindered the widespread application of these technologies. The advent of the Internet and the subsequent explosion in data accumulation catalyzed the rapid advancement of artificial intelligence content generation technology. Around 2010, researchers began utilizing deep learning techniques to generate more intricate text and multimedia content. In 2014, Google presented a seminal paper titled "Generating Images through Neural Networks," introducing the concept of Generative Adversarial Networks (GANs). As a groundbreaking technology capable of producing realistic images, GANs served as a significant milestone in the development of AIGC.

The development of Generative Language Models (GLMs) has played a crucial role in further enhancing the technology and products of AIGC. By continuously exploring and experimenting with GLMs in research and application, AIGC strives to introduce more advanced and efficient technologies, enabling it to sustain its leadership position in artificial intelligence. This chapter will delve into the market, policy, and research status of AIGC technology in China, offering comprehensive insights into its development.

\subsection{Market Status in China}

The White Paper on the Development of China's Digital Economy was recently released by the China Academy of Information and Communication Technology. According to the paper, the digital economy in China is experiencing a significant acceleration, defying the prevailing trend. In 2020, China's digital economy reached a scale of 39.2 trillion Yuan, marking an increase of 3.3 trillion Yuan compared to the previous year. This accounted for 38.6\% of the country's GDP, demonstrating a year-on-year increase of 2.4 percentage points. This development has effectively supported efforts in epidemic control, as well as economic and social progress (\cite{wp20}).

The digital economy has been rapidly developing across various regions. Taking into account the larger context, in 2020, there were 13 provinces in China with a digital economy scale exceeding 1 trillion Yuan. Furthermore, in terms of proportion to GDP, Beijing and Shanghai led the nation in digital economy advancement. Meanwhile, Guizhou, Chongqing, and Fujian exhibited the highest growth rates. While the digital economy is swiftly expanding, it continues to radiate and drive the development of surrounding regions, forming distinct models such as polar core, point axis, and multi-polar networks (\cite{cctv20}).

By the end of 2022, AIGC experienced an explosive trend in China. Advancements in data, algorithms, and computing power have led to the continuous improvement of AI technology, meeting diverse demands in areas such as text, video, and audio. It is projected that by 2025, the market size of AIGC in China will reach 777 billion Yuan. Additionally, the technology of virtual humans based on deep synthesis will also undergo growth, with its market size reaching 640.27 billion Yuan (\cite{aigc23}).

\subsection{Policy Status in China}

Currently speaking, China's leading industrial policy on AIGC mainly focuses on the combination of virtual and real industries, and the government has introduced multiple laws and regulations on the development of AIGC technology into physical industries. Table \ref{policy} shows part of China's current policies, and it can be observed that China's current focus on AIGC technology management is to prevent industrial hollowing out , and to guide cooperation between the virtual economy and the real economy (\cite{cctv20}).

\begin{table}[H]
\centering
\renewcommand{\tablename}{Table}
\caption{Some current Chinese national policies on AIGC}
\scalebox{0.8}{
\begin{tabular}{|c|c|}
\hline
Name of policy  & Contents \\ \hline
{\begin{tabular}[c]{@{}c@{}} Action Plan for the Integration \\ and Development of VR and Industry\\ Applications(2016-2022) \end{tabular}} & {\begin{tabular}[c]{@{}c@{}}By 2026, AIGC will achieve large-scale application\\ The total industrial scale exceeds 350 billion yuan\\ Cultivate 100 backbone enterprises\\ Build 10 influential AIGC  development zones  \end{tabular}} \\ \hline

{\begin{tabular}[c]{@{}c@{}} Guiding Opinions on Accelerating Scenario  \\Innovation and  Promoting High Quality \\ Economic Development through \\ High Level Application of AI \end{tabular}} & {\begin{tabular}[c]{@{}c@{}} Promote the integration of AI and real economy \\  Acceleration of AI products \\ Manage technological breakthroughs \\ Explore new paths for the development of AI \\ Using AI to promote high-quality economic development
\end{tabular}} \\ \hline

{\begin{tabular}[c]{@{}c@{}} The 14th Five Year Plan \\ for the Development of Digital Economy \end{tabular}} & {\begin{tabular}[c]{@{}c@{}} Strengthen AI Tech convergence \\ Expand AI applications \\ Enhance the consumer experience \\ Building new digital consumption scenarios
 \end{tabular}} \\ \hline

{\begin{tabular}[c]{@{}c@{}} 2022 Chinese Government Work Report \end{tabular}} & {\begin{tabular}[c]{@{}c@{}} Promote the development of the digital economy \\ Improve the governance of the digital economy \\ Improve application capabilities \\ Helping economic development\end{tabular}} \\ \hline

\end{tabular}
}
\label{policy}
\end{table}

Upon recognizing the potential risks associated with ChatGPT, various local governments acknowledged the need for regulatory measures within the AIGC industry. In this context, China took the initiative by issuing the first regulatory document, known as the Administrative Measures for Generative AI Services (Draft), which was made available for public opinion on April 11. Comprising 21 articles, the Draft outlines comprehensive provisions pertaining to the generative AI sector, encompassing aspects such as definitions, access qualifications, responsibilities and obligations, as well as punishment measures (\cite{pp23}).

China's governmental support extends to the independent innovation, promotion, application, and international cooperation concerning fundamental technologies like  algorithms or frameworks. The draft specifically advocates for the prioritization of secure and trustworthy software, tools, computing capabilities, and data resources. The supportive stance and encouragement towards the generative AI industry are explicitly articulated in the draft document, which serves as an invitation for public input.

Additionally, numerous prominent domestic technology giants, including Baidu, Tencent, Alibaba, Huawei, JD, 360, have unveiled their corresponding strategies and plans within the generative AI domain. Concurrently, several countries such as the United States, Italy, and Germany have taken note of the risks posed by ChatGPT and are contemplating the introduction of corresponding regulatory measures.

\subsection{Development Status in China}

This study offers empirical evidence to facilitate a comprehensive analysis of the current research on AIGC technology. Furthermore, the Chinese knowledge infrastructure (CNKI) database is used to examine the status of Chinese research. This paper enables an investigation of notable differences between these two research contexts.

As of June 12, 2023, 358 papers relevant to AIGC have been retrieved from the CNKI database. Most of the papers are published after March 2023, which presents a booming tendency. Due to the limited number of papers, the research field are widely separated.

\begin{figure}[H]
    \centering
    \includegraphics[scale=0.15]{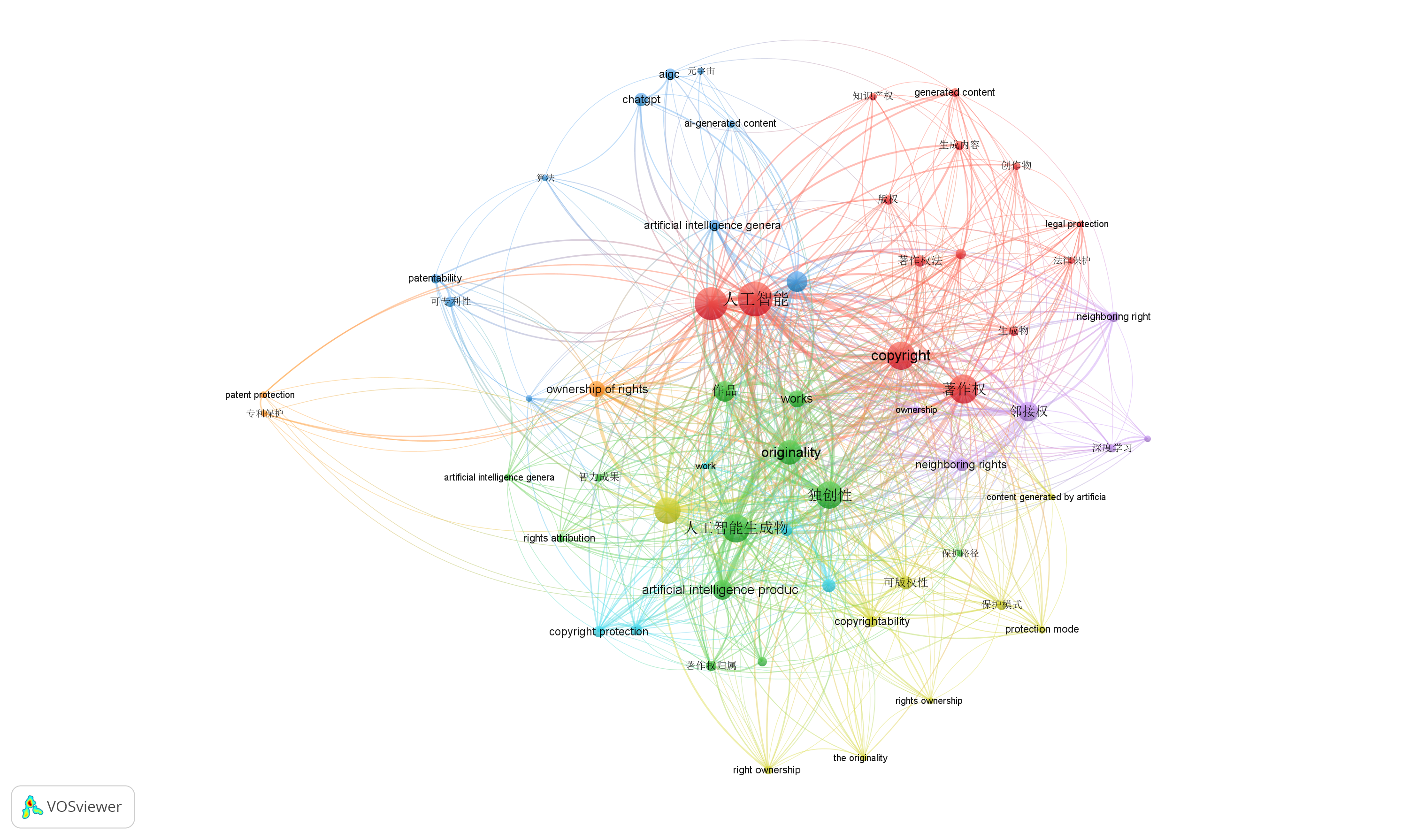}
    \caption{AIGC paper keywords}
    \label{aigcpaper}
\end{figure}

According to figure \ref{aigcpaper}, the prominent keywords associated with AIGC include AI, copyright, AI products, and originality. Analyzing the literature on AIGC published in China, it can be classified into four main categories.

The first category pertains to theoretical research, which primarily focuses on the development of AIGC. For instance, \cite{lby23} outlined the fundamental aspects for AIGC development, including network form evolution, content production, human-machine interaction mode, and network resource organization. Their investigation emphasized three dimensions: data assignment, model intelligence, and spatial empowerment, exploring the technical characteristics, elements, and developmental stages of AIGC. \cite{czf23} introduced the key aspects of AIGC technology, encompassing its historical development, technological evolution, scenarios of application, and governance. Additionally, \cite{cyw23} delved into the challenges posed by AIGC, such as unemployment, income distribution, competition, intellectual property, security, privacy, morals, ethics, energy, and the environment.

The second category involves application research, focusing on potential fields where AIGC technology can be applied. For example, \cite{ljf23} employed large language models like ChatGPT2 to train a language model capable of automatically generating traditional Chinese ancient poetry. The model performed well and successfully passed the Turing test. \cite{zy23} developed a comprehensive framework for a smart library system that integrates AIGC technology across five dimensions: infrastructure, algorithm support, data resources, business functions, and service applications. They explored the primary challenges encountered during the iterative innovation of smart libraries. \cite{wh23} discussed the prospective applications and challenges of generative artificial intelligence's big models in the Metaverse life, using medicine as an illustrative example. They examined concepts like digital biological cells and establishing connections between digital cells and neurons to enable the Metaverse to replicate the perception and biochemical reactions of the physical world for medical advancements (\cite{cheng2023metaverse}).

The third category revolves around media research, primarily investigating the relationship between AIGC and its impact on media. \cite{wj} conducted a study using BERT-DTM thematic analysis, social network analysis, and LIWC sentiment analysis methods to uncover network public opinion trends surrounding the phenomenal popularity of AIGC. The research revealed that public attention toward AIGC revolved around six major themes: technological progress, event marketing, digital art, market investment, virtual anchors, and prospects. They identified key figures in the communication network triggered by AIGC as well-known bloggers with expertise in the field. Furthermore, AIGC's emergence generated concerns about machine substitution for humans. \cite{zmn23} explored the influence of AIGC technology on the news industry, highlighting challenges such as false news dissemination and copyright disputes. They argued that the operational logic of the news industry will undergo significant changes, leading to industry restructuring and the birth of new news products and formats, promoting more personalized news consumption. \cite{mzy23} classified AIGC false information into factual and hallucinatory types based on their generation mechanism and manifestation. Their analysis covered topics like data error, author's work error, objective fact error, programming code error, machine translation error, false news events, false academic information, false health information, prejudice, and discrimination. This comprehensive examination provided a theoretical foundation for further research on AIGC false information.

The fourth category pertains to copyright issues related to AIGC, which has become an extensively researched topic in China. \cite{wq17} investigated the nature of content generated by artificial intelligence within copyright law. They argued that AIGC content, being the outcome of algorithms, rules, and templates, lacks the unique personality of human creators and cannot be recognized as a work. \cite{xmy18} proposed including AIGC works within the framework of related rights protection, presenting a solution that clarifies the distinction between human works and artificial intelligence creations while addressing the lack of legal protection for AIGC works. \cite{whd19} identified three challenges posed by AIGC technology to patent law. First, the issue of patenting AIGC works is highly debated, but granting patent protection seems to align with global trends. Second, defining AI patent subjects presents a challenge due to the "human inventor centrism" principle in current patent laws. Future laws should acknowledge machine inventors as legitimate subjects. Finally, authorizing AIGC patents is difficult, necessitating adjustments to the conditions for patent authorization to facilitate better access to patents for AIGC works.

\section{Construction of AIGC Ecology}

The advancement of AIGC has led to the appearance of the AIGC ecology. In today's dynamic world, the pursuit of ecological research on AIGC has become a priority. This research can be categorized into upstream, midstream, and downstream segments based on the product structure. Researchers are exploring ways to efficiently allocate crucial components of AIGC's ecological power, thereby facilitating its rapid development. The inclusion of integrated artificial intelligence, meta-universe, or decentralized Web 3.0 will result in varying user experiences. Connecting AIGC to the network and then to smart devices at home can potentially enhance our daily tool usage. This analysis delves into the development history of AIGC ecology in China.

\subsection{Upstream industry}

Understanding the upstream industry of AIGC is crucial as it forms the groundwork for enhancing the AIGC ecology and building the middle and downstream sectors. The establishment of the upstream industry is a fundamental part of the AIGC ecology and acts as a solid foundation for its future development. This section will explore the key factors involved in upstream molding, which is essential knowledge for comprehending the upstream work. In this context, upstream refers to the front-end industry encompassing AIGC software, hardware, personnel training, and data support.

\subsubsection{Hardware aspect}

Fast computing software is indispensable for AIGC and relies heavily on robust hardware support. The development of ChatGPT, for instance, utilized the cutting-edge A100 graphics card from NVIDIA, with a total of 10,000 units employed. This generative AI software not only necessitates access to billions of data points but also demands substantial computing power as its backbone. However, it is important to note that this is just the initial phase of ChatGPT's development.

Implicit in the increasingly pervasive use of GPUs (graphics processing units) lies their essentiality in AIGC. Originally designed for graphics processing, GPUs have found extensive application in artificial intelligence acceleration. GPU applications have proliferated in diverse areas such as generative AI, large language models, autonomous driving systems, and machine networks. Upon acquainting ourselves with ChatGPT, it becomes clear that Microsoft invested 1 billion dollars in constructing a supercomputer to power it—this laid the necessary groundwork for computational and data processing. 

Presently, the AIGC industry faces limitations in terms of hardware strength and chip computing power. China's chip sector is hindered by restrictions imposed by the United States, resulting in limited access to chips, consequently impeding the pace of AIGC industry development. Thankfully, major enterprises have made significant breakthroughs in this field, and the country is heavily investing in research and development, fostering optimism for the improvement of China's chip industry. 

\subsubsection{Personnel training}

The tech industry unequivocally hinges on talent acquisition. As science and technology advance, there is an increasing demand for skilled professionals. The state places great value on AI talent training, offering relevant programs and courses in both universities and research institutions. Scientific research projects also actively promote talent cultivation in this field. Enterprises too recognize the importance of talent acquisition as they vie for a share in the AIGC industry, with notable examples such as Huawei, Ali, Tencent, and Baidu. 

Strengthening personnel training is not restricted to augmenting higher education. As society progresses, many scientific research achievements emerge through collaborative team efforts or organizational endeavors. Nobel Prize laureates exemplify this phenomenon, as research complexity necessitates teamwork. Therefore, nurturing comprehensive qualities beyond academic training, such as social skills, communication skills, and critical thinking, is imperative.

\subsubsection{Data support}

Data support encompasses various aspects including data analysis, data processing, and data storage. The development of quality AI generation software critically relies on robust data support. The rise of big data and artificial intelligence has propelled the growth of the data processing industry. For instance, in 2019, the market size of the data annotation industry in China reached 3.09 billion yuan, and by 2020, it exceeded 3.6 billion yuan. Estimates anticipate the market size to surpass 10 billion yuan by 2025.

Once data is processed, it must be stored in suitable mediums. The concept of databases emerged in the 1970s and has since gained prominence. A database entails organized storage of data within a computer system. Compared to traditional flat files, databases offer benefits such as data security, independence, recoverability, and centralized data management. While various versions of database storage exist, they all revolve around the core concept of a data model, which represents the organization of the database in a computer-readable format.

Internationally, leading cloud data warehouse provider Snowflake witnessed a 67\% revenue increase in the third quarter of 2022 compared to the same period in the previous year, amounting to 550 million dollars. Domestically, the market size of China's big data platform surged to 5.42 billion yuan in the first half of 2021, showing a growth rate of 43.5\%. Huawei, Tencent Cloud, Ali Cloud, Baidu, and Star Ring are among the major domestic big data enterprises that have launched cloud-native data lake and cloud-native data platform products. Moving ahead, the domestic data storage industry is poised to make breakthroughs in areas like data management, data integration, and product innovation, thereby providing robust data storage support for AIGC technology (\cite{ref4131}).

\subsection{Midstream industry}

During the midstream stage of AIGC technology development, particular emphasis is placed on key factors such as cloud computing, big data, and machine learning from data. This phase is critical for the establishment of AIGC software, with China emerging as a global leader in industries such as cloud computing and big data.

Regarding cloud computing, China initially lagged behind other countries but has made significant progress in recent years. Following the national strategy, localities have intensified efforts to develop the cloud computing industry. Over 20 cities have designated cloud computing as a priority area for development and have issued industrial development plans and action plans accordingly. Examples include Beijing's "Xiangyun" plan, Shanghai's "Sea of Clouds" plan, Shenzhen's "Kun Yun" plan, Chongqing's "Cloud" plan, Ningbo's "Nebula" plan, Wuxi's "Cloud Valley" plan, Suzhou's "Colorful Cloud" plan, Harbin's "Cloud Flying" plan, Huizhou's "Huiyun" plan, Guangzhou's "Sky Cloud" plan, Inner Mongolia's "Blue Sky and White Cloud" plan, et al. (\cite{ref421}).

The construction of the computing industry's ecological chain is becoming increasingly refined. Under government supervision, cloud computing service providers, along with hardware and software suppliers, network service providers, cloud computing consulting and planning firms, delivery and maintenance services, integration service providers, and terminal equipment manufacturers, collectively form the industrial ecosystem of cloud computing.

Big data, characterized by its vast volume, is expanding exponentially. It is estimated that approximately 2.3 trillion gigabytes of data are generated daily, with this number only projected to rise. The proliferation of mobile phone networks contributes significantly to this growth. For instance, six out of the world's seven billion people now own a mobile phone, resulting in a surge in data volume from text and Whats App messages, photos, videos, and numerous applications. As businesses expand rapidly, the demand for new database management systems and IT professionals also escalates. The influx of big data is expected to create millions of new IT jobs in the coming years.

At this stage, machine model learning encompasses data processing and hardware capabilities, highlighting the significance of machine learning algorithms. The manner in which data is input, processed, and transformed into valuable and desirable content becomes crucial. While single CPU processing is effective, it often proves insufficiently fast for content production, necessitating the development of multi-threaded machine models. Currently, we have achieved the desired level of progress in this area.

\subsection{Downstream industry}

Downstream applications primarily target end consumers and require a heightened focus on software functionality and user experience. AIGC technology serves as the foundation for developing application software in various sectors, such as digital health, Internet of Things (IoT), and smart living. Furthermore, as science and technology continue to advance, AIGC applications will extend to industries like finance, healthcare, education, and media.

\subsubsection{Terminal product}

Presently, our terminal products cater to different user groups. Consumer-oriented products include Wenxin Yiyan, while enterprise-oriented products include Huawei cloud Pangu large model. As AIGC progresses, it enables effective cost reduction by minimizing labor and time expenses. AI also enhances user experience by optimizing content generation based on user behavior and feedback, leading to improved satisfaction levels.

In the field of digital medicine, a comprehensive system can be established that encompasses online search, consultations, offline visits, prescriptions, and medication processes, all accessible through personalized logins linked to individuals' real names. By leveraging global data comparisons of similar medical cases, solutions can be provided. In the absence of identical cases, attending physicians can collaborate with top international medical experts to devise suitable treatment plans. Capitalizing on its technical advantages, digital medicine helps address information asymmetry between doctors and patients, streamlines the healthcare process, reduces costs, enhances the medical experience, and improves disease diagnosis efficiency and patient management. Telemedicine services facilitated through platforms or smart devices enable more patients to access effective treatments, catering to diverse patient needs and mitigating the issue of imbalanced medical resources.

\subsubsection{Internet of Things and Web 3.0}

The Internet of Things (IoT) gained significant attention a few years ago, and with the introduction of 5G technology, the concept of IoT has gained further momentum. The idea behind IoT is to connect various objects to the network. Chinese technology companies have also collaborated with home appliance manufacturers to establish their own IoT ecosystems. For instance, Huawei's smart life ecosystem incorporates Huawei's electronic products along with other companies' appliances such as air conditioners, refrigerators, and TVs. Additionally, the cost-effective Mijia ecosystem offers a diverse range of products including mobile phones, tablets, watches, air fryers, refrigerators, and washing machines. Our focus extends beyond the mobile phone industry to embrace the concept of an ecosystem, which aligns well with the AIGC ecosystem. AIGC technology can be integrated into the IoT ecosystem and customized for each household based on their unique lifestyle and work schedules, thereby enhancing consumers' quality of life (\cite{cheng2022roadmap,huang2023roadmap}).

AIGC technology is closely intertwined with the metaverse and Web 3.0. It facilitates the rapid generation of rich and diverse content for the metaverse while enabling the development of intelligent non-player characters (NPCs) and smart contracts. Moreover, AIGC assists decentralized applications in generating and processing data, making automated decisions, and providing robust support for the advancement of the metaverse and Web 3.0. This integration enhances the realism and interest of virtual environments, and promotes decentralization and intelligence within Web 3.0 (\cite{fan2023}).

With the rapid development of AIGC, a multitude of downstream applications have emerged. AIGC is poised to become the cornerstone of internet content infrastructure, enhancing communication and enabling more efficient access to information (\cite{Cheng2016}). Currently, AIGC technology has begun penetrating highly digitized industries such as media, e-commerce, and film and television, yielding significant developmental achievements. As the market expands, several major search engines have already incorporated AIGC software. For instance, Microsoft's Bing has integrated ChatGPT, while Baidu's Wenxin Yi plans to connect it with Baidu search in the near future, enabling users to seamlessly leverage AIGC technology. The integration of AIGC and the metaverse has been proposed as engineers tackle the monumental task of building metaverse projects, which require extensive time and data resources. AIGC offers a viable solution by automatically generating characters and scenes based on inputted data, meeting our specific requirements. Web 3.0 allows for local data storage, safeguarding consumer rights and mitigating concerns such as power outages or data loss at centralized data centers. Our data remains in our possession, and even if issues arise, AIGC technology can still provide assistance, greatly facilitating our daily lives (\cite{huang2023high}).

\section{Current challenges}

Currently, AIGC faces various external challenges. The first challenge pertains to consumer mindset, as some individuals may have reservations about accepting AIGC as substitutes for human labor. This issue can stem from conflicts of interest between companies and their employees, with concerns regarding potential job displacement by AI. Prominent figures in Silicon Valley have expressed opposition to the development of artificial intelligence, highlighting valid concerns considering the uncertainty surrounding the source and future trajectory of AI advancements. Government intervention is crucial in establishing regulations and facilitating smooth business operations in this regard.

Technical limitations represent another challenge in the field of AIGC. Robust AIGC technology necessitates substantial amounts of data, efficient algorithms, and powerful computing capabilities. While notable progress has been made in certain areas, many domains still exhibit technical constraints in terms of algorithm accuracy and efficiency, particularly in image recognition, natural language processing, and machine learning. Addressing these technical challenges requires ongoing efforts from scientists to develop more efficient algorithms and enhance computational power.

Funding poses a significant challenge to the advancement of AIGC technology. Extensive financial support is required, especially during the research and development phase. However, many AIGC companies currently face funding shortages, which may impede their ability to fully develop and commercialize their technologies. To overcome this obstacle, governments and private investors are actively working to provide increased financial backing for initiatives. Moreover, some AIGC companies seek additional funding through self-financing or strategic partnerships with other organizations.

A shortage of talent represents yet another challenge for AIGC technology. The development of AIGC relies on a substantial number of highly skilled professionals, including data scientists, algorithm engineers, and computer scientists. Unfortunately, there is a global scarcity of AIGC talent. To mitigate this issue, many AIGC companies are striving to attract and nurture top-tier talent. Additionally, governments and educational institutions are making concerted efforts to cultivate a greater pool of AIGC professionals.

\begin{figure}[H]
    \centering
    \includegraphics[scale=0.6]{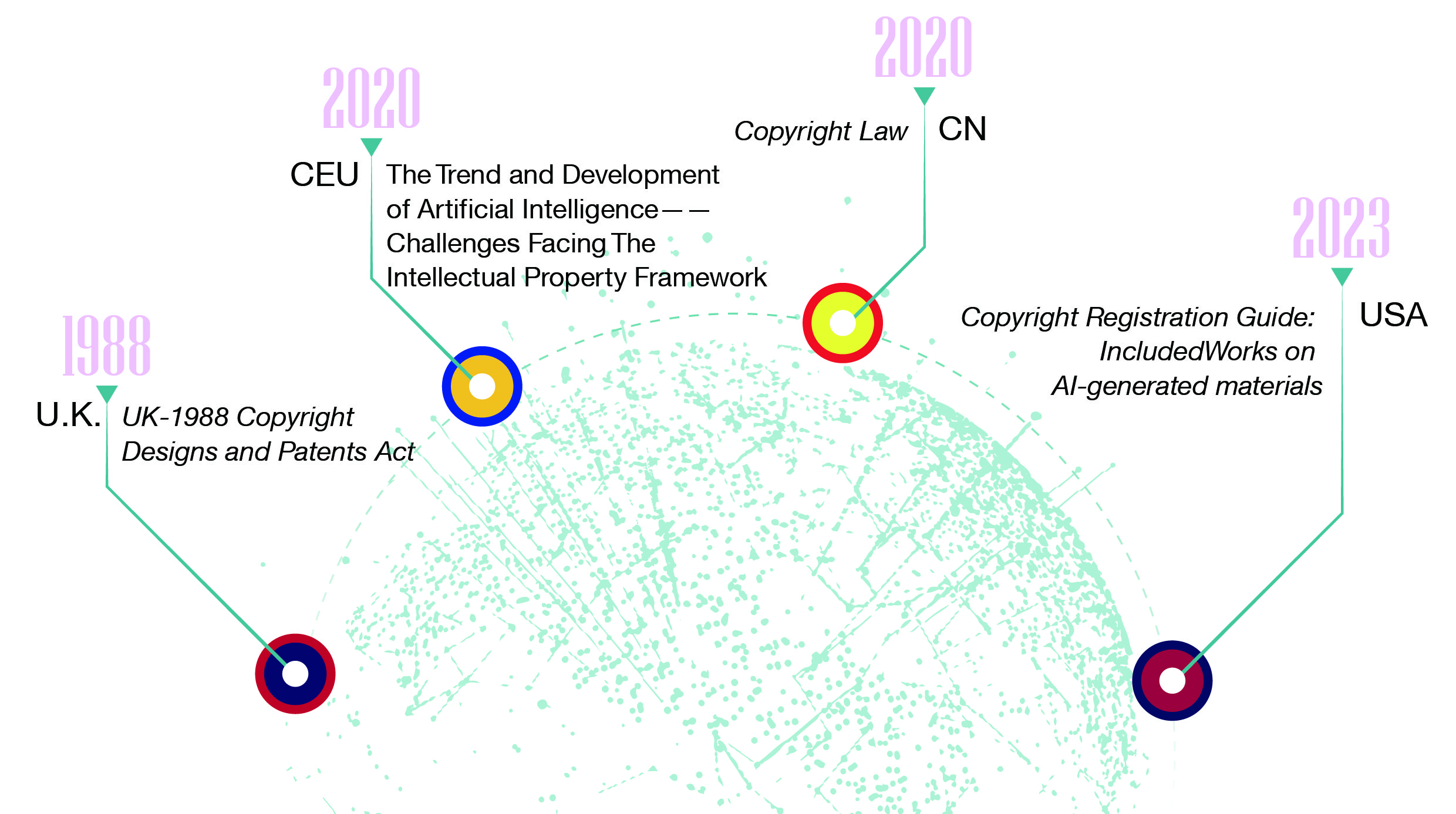}
    \caption{AIGC Legislation from different countries}
    \label{5.12}
\end{figure}

Content generated by AIGC encompasses diverse formats such as music, images, videos, and articles. The automated creation of such content raises concerns related to copyright infringement and attribution. Different countries have introduced varying legal frameworks to regulate the content produced by AIGC, aiming to address copyright issues and clarify ownership rights, as figure \ref{5.12} gives out a clear framework.

As time goes, these challenges can be overcome through ongoing research and technological advancements. Looking ahead, the prospect of AIGC holds great promise. The focus lies in the continued development of AIGC technology, extending beyond mere verbal responses. In terms of future trends, the advancement of AIGC is expected to manifest in several key aspects. Firstly, there will be an increased emphasis on enhanced intelligence, enabling AIGC systems to independently process data, learn, and make decisions autonomously. Secondly, cloud computing will play a crucial role by providing greater computational and storage capacities to support AIGC applications. This higher degree of cloud integration will empower AIGC with more robust capabilities. Lastly, AIGC will expand its reach into a wider array of industries, offering intelligent solutions tailored to the specific needs of various sectors. This diversification will deepen the integration of AIGC across different domains, fostering intelligent transformations in multiple industries.

\section{Conclusion} 
AIGC is still in its early stages of development and has enormous potential for development. With the development of science and technology, AIGC will make breakthroughs in the following two areas in the future. We drew 6 points to conclude the future development of AIGC.

Primarily, AI plays a crucial role in pre-processing and enhancing data. This involves employing techniques such as data cleaning and data enhancement to improve the quality and richness of the data. By mitigating the impact of data deviations on model training, AI assists in obtaining the necessary data for analysis and decision-making.

Secondly, it is imperative to construct interpretable models by incorporating approaches like logistic regression and decision trees. These interpretable machine learning models are designed to meet the diverse requirements of different application scenarios. By emphasizing interpretability, the aim is to facilitate wider adoption of AIGC technology, enabling more enterprises to benefit from its capabilities.

Next, enhancing the generalization ability of models is crucial. Techniques like regularization and comprehensive learning are employed to address overfitting issues, thereby improving the model's ability to generalize and increasing its reliability for practical applications.

Leveraging distributed computing resources, such as cloud computing and GPU, is another significant aspect. By harnessing these resources, the efficiency of model training and reasoning can be enhanced while reducing the computational resource costs associated with AIGC.

The fusion of multi-modal data is of great significance. Incorporating various types of data, including images, text, and speech, enables cross-modal information fusion and interaction. This approach amplifies the performance of the model and enhances its application value by leveraging complementary information from different modalities.

Lastly, ensuring privacy and security during model training is paramount. Safeguarding user and enterprise data from leakage and misuse requires stringent measures to protect against potential breaches or unauthorized access.

Through this approach, China can shape a distinctive model for AIGC development with supervision and legislation that better safeguards user safety and privacy and maintains national security. As such, it is arguable that China possesses relative strategic advantages given its unique geopolitical position, significant technological capacity, and ambitions toward technological leadership. Overall, exploring avenues for effective governance frameworks that balance innovation and business opportunities with sustainable risk management will be critical to successfully realizing AIGC's potential.

\newpage 
\printbibliography[title=References]
\end{document}